\documentclass[10pt,twocolumn,letterpaper]{article}

\usepackage[pagenumbers]{iccv} 
\usepackage{multirow}
\usepackage{enumitem}
\usepackage{amssymb}
\usepackage{svg}
\definecolor{iccvblue}{rgb}{0.21,0.49,0.74}
\usepackage[pagebackref,breaklinks,colorlinks,allcolors=iccvblue]{hyperref}

\def\paperID{*****} 
\def\confName{ICCV}
\def\confYear{2025}

\title{Exploring Spatial-Temporal Dynamics in Event-based Facial Micro-Expression Analysis}

\author{
Nicolas Mastropasqua$^{1,3}$\thanks{These authors contributed equally to this work.} \quad 
Ignacio Bugueno-Cordova$^{2,4}$\footnotemark[1] \quad 
Rodrigo Verschae$^{2}$ \quad 
Daniel Acevedo$^{1,3}$ \\ 
Pablo Negri$^{1,3}$ \quad 
Maria E. Buemi$^{1,3}$  \bigskip \\ 
$^1$Universidad de Buenos Aires, Facultad de Ciencias Exactas y Naturales, Argentina \\
$^2$Institute of Engineering Sciences, Universidad de O'Higgins, Chile\\
$^3$CONICET-UBA, Instituto de Ciencias de la Computacion (ICC), Argentina \\
$^4$L3S Research Center, Leibniz Universität Hannover, Germany\\
{\tt\small\{nmastropasqua, dacevedo, pnegri, mebuemi\}@dc.uba.ar}\\
{\tt\small i.bugueno@ieee.org, rodrigo@verschae.org}
}

\begin{document}
\maketitle
\begin{abstract} 

Micro-expression analysis has applications in domains such as Human-Robot Interaction and Driver Monitoring Systems. Accurately capturing subtle and fast facial movements remains difficult when relying solely on RGB cameras, due to limitations in temporal resolution and sensitivity to motion blur. Event cameras offer an alternative, with microsecond-level precision, high dynamic range, and low latency. However, public datasets featuring event-based recordings of Action Units are still scarce.
In this work, we introduce a novel, preliminary multi-resolution and multi-modal micro-expression dataset recorded with synchronized RGB and event cameras under variable lighting conditions. 
Two baseline tasks are evaluated to explore the spatial-temporal dynamics of micro-expressions: Action Unit classification using Spiking Neural Networks (51.23\% accuracy with events vs. 23.12\% with RGB), and frame reconstruction using Conditional Variational Autoencoders, achieving SSIM = 0.8513 and PSNR = 26.89 dB with high-resolution event input.
These promising results show that event-based data can be used for micro-expression recognition and frame reconstruction. 

\end{abstract}    
\section{Introduction}
\label{sec:intro}

Sensing and analyzing micro-expressions provides information about subtle human emotions. Facial Expression Recognition (FER) within Affective Computing has applications in social robotics~\cite{Ramis2020UsingAS} and intelligent vehicles, where Driver Monitoring Systems (DMS) use facial expressions—both macro~\cite{Xiang2024AMD, Chen2020EDDDED} and micro~\cite{Vijay2021RealTimeDD}—to detect drowsiness and emotional states, supporting road safety~\cite{Yang2023AIDEAV}. Micro-expression analysis also contributes to Autism Spectrum Disorder detection~\cite{Li2019AFA}.

\begin{figure}[t]
  \centering
   \includegraphics[width=\linewidth,trim={0 400 0 0},clip]{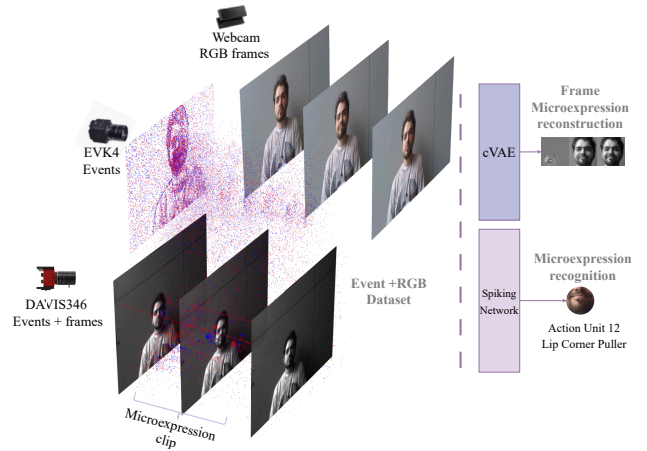}
   \caption{Our proposed preliminary dataset to study Action Unit recognition and Facial micro-expression reconstruction consists of multi-modal (RGB and Events) and multi-resolution recordings of micro-expressions using a DAVIS346 event camera, a Prophesee EVK4 event camera, and an ASUS ROG Eye S RGB camera.}
   \label{fig:onecol}
\end{figure}

Deep learning models trained on large FER datasets achieve high performance for posed expressions representing Ekman's six basic emotions~\cite{ekman} under controlled conditions. However, performance decreases in 'in-the-wild' settings with varied lighting, occlusions, and poses~\cite{Kopalidis2024AdvancesIF}. Categorizing emotions into a few archetypal expressions oversimplifies their diversity~\cite{Plutchik1980AGP, TenHouten2021BasicET}, introducing label ambiguity and inconsistency across datasets~\cite{Zeng2018FacialER}.

An alternative and more descriptive codification is given by the Facial Action Coding System (FACS)~\cite{Ekman1978FacialAC}, which aims to identify precise facial micro-movements called Action Units (AUs). AU-labeled datasets remain limited due to the time-intensive, expertise-dependent annotation process. Moreover, RGB cameras impose temporal constraints, as micro-expressions last between 170–500 ms, with a lower bound near the frame rate of standard cameras~\cite{Yan2013HowFA}. As a result, current sensors and methods remain insufficiently robust for AU recognition~\cite{Li2021DeepLF}.
Alternative sensing modalities, such as thermal~\cite{Adra2024BeyondRT} and event cameras~\cite{Becattini2024NeuromorphicFA}, address these limitations. 

Event cameras provide microsecond-level temporal resolution, low latency, high dynamic range ($\geq$ 120 dB), and reduced power consumption. They also improve privacy, as re-identification from event data is more challenging than from RGB images~\cite{Ahmad2023PersonRW}. Although there has been an increased interest in using event cameras for facial analysis~\cite{Becattini2024NeuromorphicFA2}, publicly available event-based micro-expression datasets remain scarce. Recent efforts include VETEX~\cite{Adra2024BeyondRT}, covering only a subset of AUs, and FACEMORPHIC~\cite{Becattini2024NeuromorphicFA}, which is not yet fully accessible. Simulated event streams from RGB videos, using tools like v2e~\cite{Hu2021v2eFV} and ESIM~\cite{Rebecq2018ESIMAO}, offer an alternative but do not replicate the temporal dynamics and noise characteristics of real neuromorphic data.

Processing data from event cameras requires models that operate natively on sparse, temporally precise signals. Spiking Neural Networks (SNNs)\cite{Maass1996NetworksOS}, inspired by biological neurons, are naturally suited for this representation. While most deep learning models use backpropagation with labeled datasets, brain-inspired unsupervised local rules, such as Spike Timing Dependent Plasticity (STDP) allow SNNs to adapt by modulating synaptic strengths based on the Hebbian principle. However, STDP alone is often insufficient for training deep SNNs, motivating the use of surrogate gradient descent to address spike non-differentiability with smoothing functions~\cite{Neftci2019SurrogateGL}.

In this work, we propose to explore the spatial-temporal dynamics of micro-expressions using event-based facial data. To this end, we introduce a preliminary multi-modal, multi-resolution dataset of performed Action Units recorded with synchronized RGB and event cameras under varying lighting conditions. Using this dataset, we evaluate two baseline tasks: (1) Action Unit recognition with Spiking Neural Networks trained via surrogate gradient descent, and (2) frame reconstruction from event data using Conditional Variational Autoencoders (cVAE), comparing low and high-resolution event inputs.

The main contributions of this work are:
\begin{itemize}
    \item A preliminary multi-modal, multi-resolution dataset of RGB and event data under mixed lighting conditions and posed Action Units. A larger version of this dataset is under construction and is expected to be publicly available upon request soon.  
    \item An exploration of Spiking Neural Networks trained with surrogate gradient descent to learn spatial-temporal dynamics for Action Unit recognition on this dataset.
    \item An exploration of intensity frame micro-expression reconstruction models conditioned on event-based inputs, highlighting differences between low and high resolution data from DAVIS346 and EVK4, respectively, in terms of reconstruction quality.
\end{itemize}

The rest of this paper is organized as follows: 
Section~\ref{sec:related-work} reviews related work; 
Section~\ref{sec:3_dataset} introduces the novel and preliminary multi-modal dataset for micro-expression analysis;
Section~\ref{sec:4_methodology} presents the proposed baselines for exploring the spatio-temporal dynamics in facial micro-expression recognition and intensity reconstruction;
Section~\ref{sec:training-details} describes the training details;
Section~\ref{sec:results} reports the preliminary results; 
and Section~\ref{sec:6_conclusions} concludes and projects future work.
\section{Related work}
\label{sec:related-work}

\subsection{Event-based micro-expression datasets}

While micro-expression datasets for the RGB modality are relatively small and scarce, their event-based counterpart is no exception. Recently, the community has made valuable efforts to address this problem in multiple ways: Berlincioni \etal~\cite{Berlincioni2023NeuromorphicEF} introduced  NEFER, a dataset for emotion recognition which consists of paired RGB and Event videos annotated with the six basic emotions, as well as landmarks and bounding boxes. However, it is not well-suited for micro-expression analysis since it does not provide AU  annotations. The first event dataset containing a large subset of Action Units to date was proposed by Becattini \etal~\cite{Becattini2024NeuromorphicFA}. FACEMORPHIC contains 64 subjects performing 24 AUs recorded with a $640 \times 480$ RGB camera and a $1280 \times 720$ event camera, which are temporally synchronized. Similarly, Adra \etal~\cite{Adra2024BeyondRT} proposed the VETEX dataset containing event, RGB and thermal recordings of 30 participants performing a subset of seven AUs under natural and controlled artificial illumination. However, to date, both of these datasets have yet to be made publicly available. 

An alternative approach to collecting real event data consists of simulating an event stream from RGB videos. For example, Verschae \etal~\cite{Verschae2023EventBasedGA} used the v2e simulator~\cite{Hu2021v2eFV} to obtain an event-based version of the CK+~\cite{Lucey2010TheEC} and MMI \cite{Pantic2005WebbasedDF} datasets,  both well-known benchmarks for posed FER and AU recognition. They showed that the Asynchronous Sparse Neural Network (Asynet)~\cite{Messikommer2020EventbasedAS} can achieve promising results on the proposed datasets. Although working with simulated events facilitates training and testing new algorithms, a significant domain shift still occurs between simulated and real event data: for example, compression artifacts in videos might lead to the generation of noisy events during the simulation~\cite{Berlincioni2024NeuromorphicVA}. Furthermore, simulated events share a series of discrete timestamps, unlike real data. This can accentuate the domain shift for models sensitive to timestamp distribution, such as Spiking Neural Networks~\cite{Zhang2023V2CEVT}.

\subsection{Spiking Neural Networks}

The neuromorphic-driven nature of Spiking Neural Networks (SNN) makes them suitable for processing event camera streams. Spiking neurons communicate with each other by sending discrete spikes when their accumulated membrane potential surpasses a threshold value. 
Moreover, existing neuromorphic hardware, such as TrueNorth~\cite{Merolla2014AMS} and Loihi~\cite{Davies2018LoihiAN}, offer exciting opportunities to leverage these features.

However, SNNs still fall short of ANNs in terms of their training capabilities, generally struggling to achieve comparable performance. Training methods for SNNs can be grouped into three categories: supervised learning with gradient descent, unsupervised learning with local learning such as Spike-Timing-Dependent Plasticity (STDP), and reinforcement learning using reward-modulated plasticity~\cite{Yamazaki2022SpikingNN}. Existing works have studied the application of SNNs on different tasks and datasets, from small complexity to more challenging ones. For example, Vasudevan \etal~\cite{Vasudevan2021SLAnimalsDVSES} employed SNNs trained with different learning algorithms based on backpropagation for gesture recognition, obtaining a reasonable baseline in their proposed SL-ANIMALS-DVS dataset and good results in the DVS-GESTURES dataset. Dong \etal~\cite{Dong2022AnUS} proposed the use of temporal batch STDP and designed an adaptive synaptic filter, together with an adaptive lateral inhibitory connection, achieving current state-of-the-art performance for unsupervised STDP-trained SNNs on MNIST and FashionMNIST. They also showed that their SNNs outperformed ANNs for smaller training sizes of MNIST. However, these techniques still struggle to train SNNs capable of performing as well as ANNs. Interestingly, state-of-the-art results are commonly achieved by converting trained ANNs to SNNs~\cite{Yamazaki2022SpikingNN}.

Currently, to the best of our knowledge, there are no works using SNNs for AU recognition on real event data.  To evaluate its viability, we consider a standard SNN
architecture with learnable membrane time constant based on the formulations given by Fang \etal~\cite{Fang2020IncorporatingLM}. 

\subsection{Event-based facial reconstruction for micro-expression analysis}

Reconstruction of facial appearance from event data supports tasks such as AU recognition and micro-expression analysis, where subtle motion cues need to be preserved. Several methods have addressed scene reconstruction from event streams. Scheerlinck \etal~\cite{scheerlinck2020} proposed an efficient network with reduced parameters and inference time. Pan \etal~\cite{pan2022} introduced the Event-based Double Integral model to recover sharp high-frame-rate videos from blurred images and events. 

For facial analysis, Barua \etal~\cite{barua2016} developed a patch-based model capable of reconstructing faces and demonstrated direct face detection from events. Zou \etal~\cite{zou2021learningtoreconstruct} proposed a convolutional recurrent network to reconstruct temporally consistent HDR videos, validated on a dedicated dataset. Recently, Verschae \etal~\cite{verschae2025evtransfer} proposed evTransFER, a framework that employs an adversarially trained encoder for facial reconstruction and reuses it for expression recognition. While the method focuses on macro-expressions, it demonstrates that reconstruction from event data can provide useful representations for facial analysis. To date, however, micro-expression reconstruction from real event streams remains unexplored.

In this work, we explore Variational Autoencoders conditioned on real event data as a baseline approach to generate reconstructed frames of micro-expressions, enabling a qualitative assessment of the temporal dynamics captured across different sensor resolutions.
\section{Dataset collection}
\label{sec:3_dataset}

To improve the public availability of event-based micro-expression data and help the community deepen the study in this area, we collected a reduced set of videos of subjects performing a subset of Action Units according to FACS. This, in turn, allowed us to assess the viability of SNNs for AU recognition in this scenario.
All videos were recorded with an ASUS RGB webcam and DAVIS346 and Prophesee EVK4 event cameras. People aged 18 to 65 years were invited to participate in the experiment.

\subsection{Annotation schema}

We specifically selected 21 AUs focusing on potential Driving Monitoring Systems applications. According to~\cite{Vural2007DrowsyDD}, some AUs are indicative of certain levels of drowsiness before falling asleep: AU 45 (blink), AU 2 (outer brow
raiser), AU 15 (frown), AU 17 (chin raise), and 9 (nose wrinkle).
Some AUs represent macro-expressions like turning the head to one side or moving the eyes. We also considered those actions, as they could be related to potential distractions while driving.
Based on EMFACS~\cite{Friesen1983EMFACS7EF}, it is possible to decode a group of AUs activations into one of the six basic emotions. The rest of the AUs were selected to complete the codification of Happiness, Anger, and Surprise.

\begin{table}[h]
  \centering
  \begin{tabular}{cc}
    \toprule
    \textbf{AU Number} & \textbf{Description} \\
    \midrule
    AU 2   & Outer Brow Raiser \\
    AU 4   & Brow Lowerer \\
    AU 5   & Upper Lid Raiser \\
    AU 6   & Cheek Raiser \\
    AU 9   & Nose Wrinkler \\
    AU 12  & Lip Corner Puller \\
    AU 15  & Lip Corner Depressor \\
    AU 17  & Chin Raiser \\
    AU 23  & Lip Tightener \\
    AU 26  & Jaw Drop \\
    AU 41  & Lid Droop \\
    AU 43  & Eyes Closed \\
    AU 45  & Blink \\
    AU 51  & Head Turn Left \\
    AU 52  & Head Turn Right \\
    AU 53  & Head Up \\
    AU 54  & Head Down \\
    AU 61  & Eyes Turn Left \\
    AU 62  & Eyes Turn Right \\
    AU 63  & Eyes Up \\
    AU 64  & Eyes Down \\
    \bottomrule
  \end{tabular}
  \caption{List of the 21 selected Action Units (AUs)~\cite{Ekman1978FacialAC} with their corresponding descriptions, chosen based on their possible relevance to drowsiness and potential distractions in driving scenarios, as well as some emotions that could affect road safety.}
  \label{tab:au_table}
\end{table}

Unlike standard FACS-coded RGB datasets based on certified human coders, our labeling process was carried out through a guided elicitation process. For each AU, subjects were shown a video demonstration of an individual performing the target AU, and then they were instructed to perform the AU at a natural speed. This method relies on the subject's ability to mimic the demonstrated expression and our visual judgment. This is to highlight that the labels represent elicited performances intended to activate specific AUs, rather than expert-coded FACS activations.

\begin{figure}[h]
  \centering

    \subcaptionbox{}[0.24\linewidth]{%
        \includegraphics[width=\linewidth]{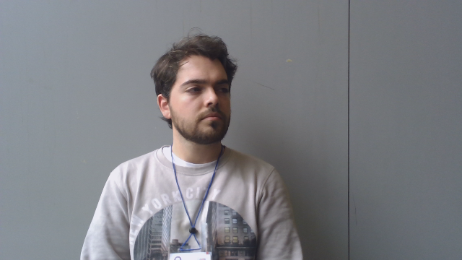}
    }
    \subcaptionbox{}[0.24\linewidth]{%
        \includegraphics[width=\linewidth,trim={0 0 0 65},clip]{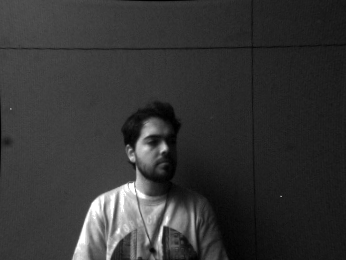}
    }
    \subcaptionbox{}[0.24\linewidth]{%
        \includegraphics[width=\linewidth,trim={0 0 0 65},clip]{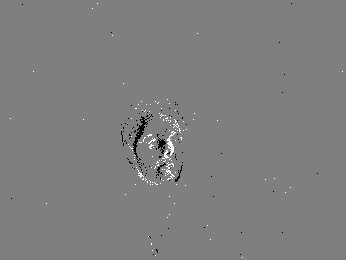}
    }
    \subcaptionbox{}[0.24\linewidth]{%
        \includegraphics[width=\linewidth]{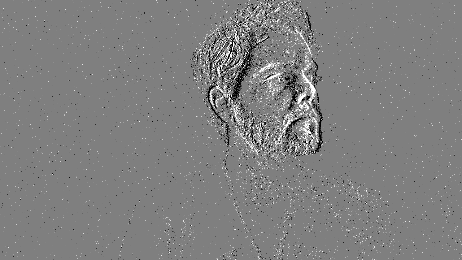}
    }
    
    \subcaptionbox{}[0.24\linewidth]{%
        \includegraphics[width=\linewidth]{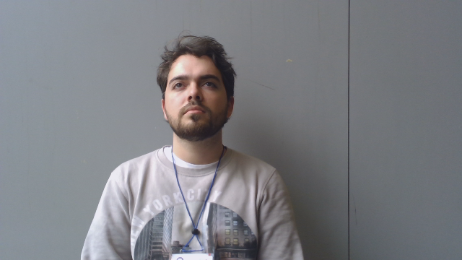}
    }
    \subcaptionbox{}[0.24\linewidth]{%
        \includegraphics[width=\linewidth,trim={0 0 0 65},clip]{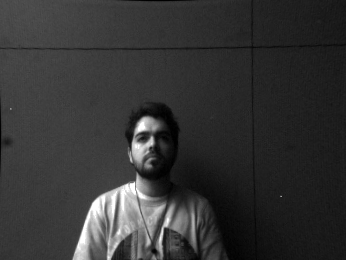}
    }
    \subcaptionbox{}[0.24\linewidth]{%
        \includegraphics[width=\linewidth,trim={0 0 0 65},clip]{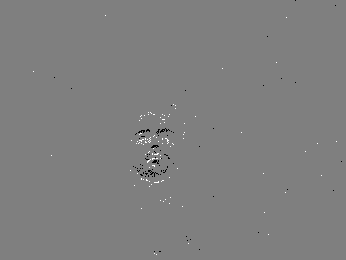}
    }
    \subcaptionbox{}[0.24\linewidth]{%
        \includegraphics[width=\linewidth]{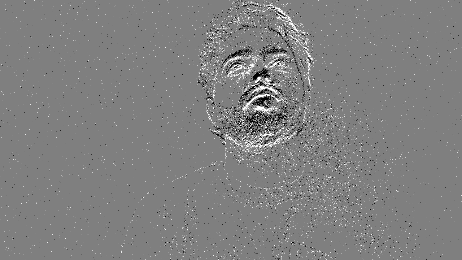}
    }

    \subcaptionbox{}[0.24\linewidth]{%
        \includegraphics[width=\linewidth]{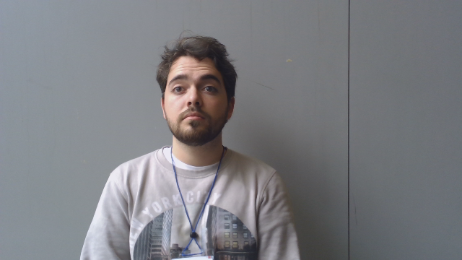}
    }
    \subcaptionbox{}[0.24\linewidth]{%
        \includegraphics[width=\linewidth,trim={0 0 0 65},clip]{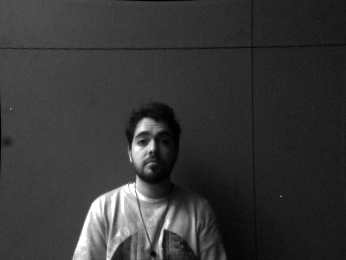}
    }
    \subcaptionbox{}[0.24\linewidth]{%
        \includegraphics[width=\linewidth,trim={0 0 0 65},clip]{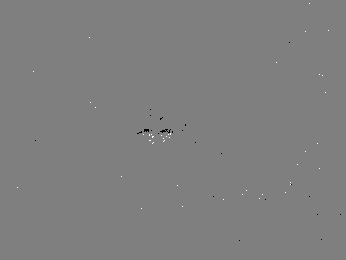}
    }
    \subcaptionbox{}[0.24\linewidth]{%
        \includegraphics[width=\linewidth]{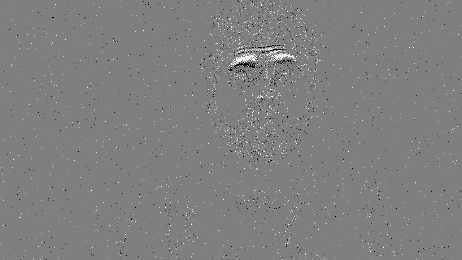}
    }

   \caption{Multimodal Facial Micro-Expression Dataset. From left to right: ASUS webcam RGB frame; DAVIS346 frame; DAVIS346 event frame; EVK4 event frame. The frames show the apex of a) AU 51 (Head Turn Left), b) AU 53 (Head Turn Up), and c) AU 2 (Outer brow raiser).}
   \label{fig:dataset}

\end{figure}

\subsection{Set-up}

The experiments were set up in an auditorium with a fully windowed wall and plenty of direct sunlight during the day.  Subjects participated in the experiment at different times of day, ranging from morning to late afternoon, resulting in non-uniform lighting conditions across recordings. Consequently, some participants were recorded with additional artificial ceiling illumination to compensate for suboptimal lighting during late afternoon. Facial illumination was measured for each subject, with lux levels ranging from 150 to 1400. Distractors might have been present during the recording since subjects were not completely isolated from other people passing by in the room.

To record each subject, we employed three cameras: a DAVIS346 event camera recording events and synchronized grayscale frames at $346 \times 260$ resolution at a maximum of 40 FPS, a Prophesee EVK4 event camera recording events at $1280 \times 720$ resolution, and an ASUS ROG Eye S RGB webcam with a resolution of $1920 \times 1080$ at a maximum of 60 FPS. All three devices were mounted on a fixed camera tripod with a special adaptation that allowed for horizontal alignment of the cameras. The cameras were connected via USB cable to a laptop, and a ROS script was used to ensure simultaneous recording and temporal synchronization. During the experiment, subjects sat comfortably in a chair in front of the cameras at a fixed distance of 80 cm, ensuring that their face was centered within the frame. Sequences of AUs were displayed on an external LCD monitor mounted on a desk adjacent to the camera tripod.
Before the experiment, both event cameras went through a calibration process to estimate their intrinsic parameters.

\subsection{Protocol}

We introduced each subject to the experimental procedure, briefly discussing the goal of our work. After the subject agreed to sign the informed consent, the experimenter described the subset of AUs to be performed by the subject, then showed them short clips with clear textual descriptions of how to perform the actions. After confirming their understanding of each AU, subjects were then instructed to sit comfortably on a chair, which was adjusted to meet the necessary distance to the cameras. Before the experiment's main run, a demonstration run was carried out to test the subject's understanding of the entire task and the timing for performing the actions.

Once the subject was ready, the main run was executed. During each run, subjects were asked to look at the screen where a video instructed them which AU to perform and when. For each AU, the video showed a clip demonstrating the action three times. A countdown with sound indicated the subject to look at the camera and be ready to perform the action. After a sufficient period, a double beep and a message on the video instructed them to rest and relax for a few seconds. Unless they warned the experimenter about a possible mistake, the next segment with the next AU played, and the subject performed the action in the same manner. The experimenter started recording each segment independently as soon as the countdown appeared on the video. If the subject or the experimenter made a mistake, the video was paused, and the action was re-done. Each main run lasted approximately 7 minutes, and each participant went through only one main run.

\subsection{Dataset description}

Our preliminary dataset consists of 7 individuals, six men and one woman, ages ranging from 18 to 27. For each of them, we selected 21 videos of elicited AU performances recorded in both RGB and Event modalities with different image resolutions, yielding a total of 510 clips with evenly distributed classes of AUs. Lighting conditions were not uniform across recordings, with lux ranging from 150 to 1400 using mainly natural illumination. A few samples obtained during our recordings can be seen in Figure~\ref{fig:dataset}, and the description of our dataset is summarized in Table~\ref{tab:dataset}.

We trimmed each clip, ensuring that only the elicited AU was present (from onset to apex and offset). In this way, we eliminated blinking and any other facial or body movements before and after the AU performance, leaving, when possible, about a second of frames of margin with no apparent activity. However, it is worth noting that some participants had difficulties isolating single Action Units. For example, some of them would move their mouth while performing AU 6 (cheek raiser) or even blink, AU 45, while doing another action. 
We decided to be less selective in this stage and keep these non-ideal clips, favoring quantity. As a consequence, some clips could contain co-activations of AU, but, in that case, only the elicited AU was labeled. The average duration of clips was 2.5s (including the safe margins of neutral activity before and after AU performance).   

As a final step, we took advantage of the synchronized frames of the DAVIS346 and extracted the bounding box positions to crop each event frame using the MTCNN detector~\cite{Xiang2017JointFD}. We note that it could also be possible to extract facial landmarks and bounding boxes from the EVK4 event frames using the synchronized stream from the ASUS RGB camera.

\begin{table}[h]
  \centering
  \begin{tabular}{ccll}
    \toprule
    \#Vids. & \#Subj. & Modality          & Resolution \\
    \midrule
     147 &  7    & RGB (ASUS)               & 1920 $\times$ 1080 \\
     147 &   7   & Events (Davis)    & 346 $\times$  260 \\
      147 &   7   & RGB (Davis)    & 346 $\times$  260 \\
     63  &   3   & Events (Evk4)     & 1280 $\times$  720 \\
    \bottomrule
  \end{tabular}
  \caption{Summary of our multi-modal, multi-resolution preliminary dataset. A total of 7 subjects were recorded, with only 3 using the complete camera setup.}
  \label{tab:dataset}
\end{table}

\section{Facial Micro-Expression Analysis Framework}
\label{sec:4_methodology}

\subsection{Main problem formulation}
Given an event stream recording from an event camera $E_{i}(t_N)=\{e_{k}\}_{k=1}^{N_i}=$
$\{(  x_ {k}, \ y_ {k},\ t_ {k},\ p_ {k} )\}_{k=1}^{N_i}$ and its label $l_i$ representing the AU performed during the $i$-th recording of our dataset, we optimize the parameters of an SNN model $F_{\theta}$ for the single-label, multi-class classification problem consisting of 21 classes of AUs present in our dataset.

\subsection{Event data representation}

Each event $e_k = (  x_ {k}, \ y_ {k},\ t_ {k},\ p_ {k} )$ is a 4-tuple representing a brightness change in pixel $(x_ {k}, \ y_ {k})$ at time $t_{k}$ above certain threshold since the previous event at that pixel. Here, $p_k\in \{-1,1\}$ indicates the sign of the brightness change.
A widely used event representation, which allows the use of traditional computer vision techniques, consists of fixing a slice of time and integrating the events in that period into a frame. Assuming a slice of size $T$ with starting and ending times $t_k$ and $t_l$, an event frame $F_p$ with two channels corresponding to polarities $p=1$ and $p = -1$ is defined as: 
\begin{equation}
F_p(x,y) = \sum_{i=t_k}^{t_l} \mathcal{I}_{p,x,y}(p_i, x_i, y_i),
\label{eq:er}
\end{equation}
where $\mathcal{I}$ is the indicator function. One drawback of this representation is that much of the asynchrony of events is lost if $T$ is too large, defying the nature of events. However, it is a straightforward way to transform the event stream into a synchronous frame representation, facilitating the use of common deep ANNs architectures.

\subsection{Baseline Architecture for AU recognition}

\subsubsection{Spiking Neurons} 

LIF spiking neurons are the essential building block of the proposed SNNs due to their low computational cost. They can be seen as a simplified bio-inspired model resembling the dynamics of an RC circuit. The subthreshold dynamics of LIF Neurons are given by the differential equation:
\begin{equation} 
\tau  \frac {dV(t)}{dt}  =-(V(t)-  V_ {rest}  )+X(t)  
\label{eq:lif}
\end{equation}
where $V(t)$ is the membrane potential at time $t$, $X(t)$ is the input to neuron at time $t$, $\tau$ is the membrane time constant and $V_{rest}$ is the resting potential. When $V(t)$ exceeds the threshold $V_{th}$, the neuron produces a spike and its membrane potential is set to $V_{reset}$. 
The discrete implementation of Eq.~\ref{eq:lif} can be derived as:
\begin{equation}
    H[t]=V[t-1]-  \frac {1}{\tau }  (V[t-1]-  V_ {rest}  )+X[t]
\label{eq:lif_discrete}
\end{equation}
The membrane time constant $\tau$ is usually a hyperparameter of the network. However, based on the Parametric Leaky Integrate-and-Fire (PLIF) unit proposed in~\cite{Fang2020IncorporatingLM}, we let $\frac{1}{t} = sigmoid(w)$ to be a learnable parameter, shared only within the neurons in the same layer. From Eq.~\ref{eq:lif_discrete} and setting $V_{reset} = 0$, it can be seen that LIF and PLIF neurons resemble the functionality of RNNs. They are both stateful; in the case of LIF, they ''remember'' the membrane potential $V(t)$; they ''forget'' according to the membrane time constant $\tau$; and they also take current input information $X(t)$ to update their next state. This highlights the potential use of PLIF SNNs to learn the spatial-temporal dynamics in micro-expressions. 

\subsubsection{Action-Unit recognition}

\begin{figure}[!t]
  \centering
   \includegraphics[width=\linewidth]{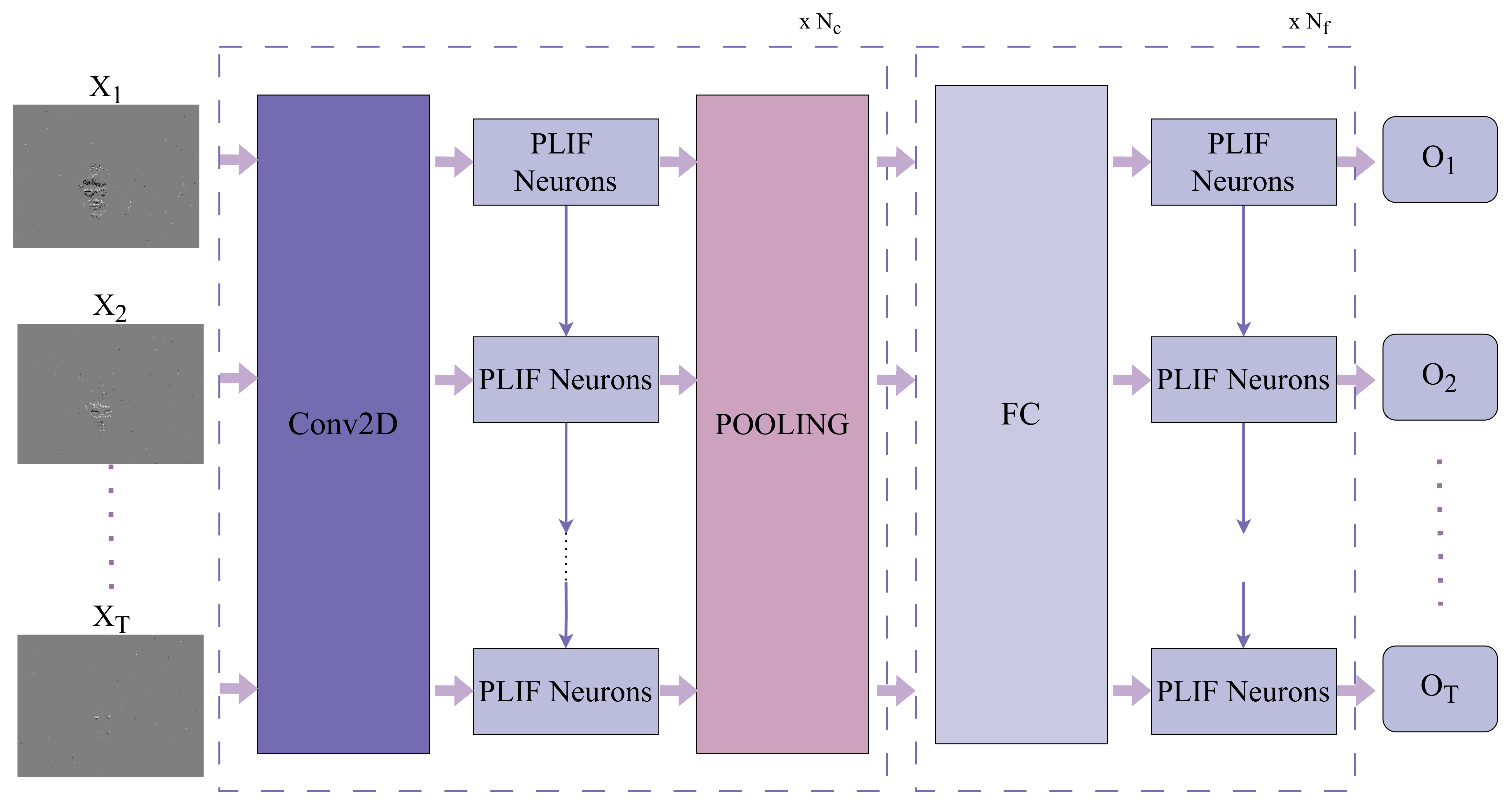}
   \caption{SNN architecture employed for AU recognition. Here, CONV2D represents a sequential block of 2D Convolution followed by Batch Normalization, and FC a sequential block of Dropout followed by a FC layer. The spiking encoder is composed of $N_c$ blocks of Conv2D-PLIFNeurons-2DMaxPooling. The classification layers FC-PLIFNeurons are repeated $N_f$ times. In this work, we set $N_c=N_f=2$.}
   \label{fig:snn}
\end{figure}

We employed a convolutional architecture with PLIFs as its building block based on the formulations given by Fang\etal~\cite{Fang2020IncorporatingLM} and represented in Figure~\ref{fig:snn}.

The proposed architecture consists of two sequential blocks of [Conv2D-BatchNorm-PLIFNeurons-MaxPool2D]. All Conv2D layers were set with a kernel size = 3, stride = 1, padding = 1, and output feature map of size 32. The 2D Pooling layers were set with kernel size = 2 and stride = 2. These blocks act as a learnable encoder to convert images to spikes. After the spiking encoder, we add two sequential blocks of [Dropout-FClayer-PLIFNeuron] to learn the classification task. The output feature sizes of the FC layers are 512 and 210, respectively. A final voting layer, implemented as an Average pooling layer of stride 10 and kernel size 10, takes the last output spikes and maps them to a final response corresponding to one of our 21 classes. When a sequence of $T$ frames is consumed, we take the average along the time dimension, yielding a vector of average activations per class.

To train the SNN using backpropagation, a surrogate gradient function is needed due to the non-derivability of the heavy-side step function that simulates the spikes. We chose arctangent as the smoothing function, and we employed Mean Squared Error (MSE) as the loss function.

\subsection{Baseline Architecture for Frame Micro-expression Reconstruction}

\begin{figure}[!t]
    \centering
    \includegraphics[width=\linewidth]{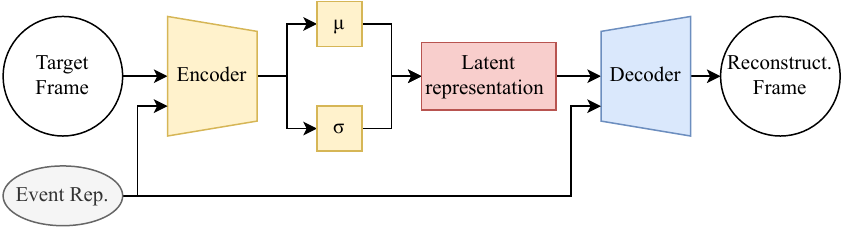}
    \caption{Conditional Variational Autoencoder architecture for face reconstruction. The model reconstructs the target frame $(x)$ conditioned on the event-based representation $(c)$ obtained from accumulated events.}
    \label{fig:face_reconstruction_cvae_architecture}
\end{figure}

For intensity frame micro-expression reconstruction, we implement a cVAE as a baseline architecture. The model conditions on an event-based representation $c$ and learns to reconstruct the corresponding ground-truth frame $x$, as shown in Figure~\ref{fig:face_reconstruction_cvae_architecture}.

Formally, the encoder $q_\phi(z \mid x, c)$ receives the target frame $x$ together with its conditional event-based representation $c$ and maps them into a latent Gaussian distribution. This distribution is parameterized by a mean $\mu_\phi(x,c)$ and a diagonal covariance given by $\sigma_\phi^2(x,c)$:
\begin{equation}
q_\phi(z \mid x, c) = \mathcal{N}\!\left(z \,\middle|\, \mu_\phi(x,c), \operatorname{diag}\big(\sigma_\phi^2(x,c)\big)\right).
\end{equation}
To enable backpropagation through the stochastic sampling process, the latent code $z$ is obtained via the reparameterization trick. In this formulation, $z$ is expressed as a deterministic function of $\mu_\phi(x,c)$, $\sigma_\phi(x,c)$, and a noise variable $\epsilon \sim \mathcal{N}(0, I)$:
\begin{equation}
z = \mu_\phi(x,c) + \sigma_\phi(x,c) \odot \epsilon.
\end{equation}
The decoder $p_\theta(x \mid z, c)$ then takes the latent vector $z$ concatenated with the conditional input $c$ and generates the reconstructed frame. The output $\hat{x}$ represents the approximation of the original target frame $x$:
\begin{equation}
\hat{x} = p_\theta(x \mid z, c).
\end{equation}
In practice, the encoder consists of two convolutional layers with ReLU activations followed by fully connected layers that produce $\mu$ and $\sigma$ with latent dimensionality 32. The decoder receives the sampled latent vector concatenated with the flattened conditional input and generates $\hat{x}$ through fully connected and transposed convolutional layers with sigmoid activation.
The network is optimized by minimizing the conditional Evidence Lower Bound (ELBO):
\begin{equation}
\begin{split}
\mathcal{L}(\theta, \phi; x, c) = &
\ \mathbb{E}_{q_\phi(z \mid x,c)} \big[ \log p_\theta(x \mid z, c) \big] \\
& - D_{\mathrm{KL}}\!\left( q_\phi(z \mid x,c) \,\|\, p(z) \right),
\end{split}
\end{equation}
with $p(z) = \mathcal{N}(0,I)$.
To evaluate the reconstruction performance, multiple image similarity metrics were calculated, including Mean Squared Error (MSE), Root Mean Squared Error (RMSE), Mean Absolute Error (MAE), Peak Signal-to-Noise Ratio (PSNR), Structural Similarity Index Measure (SSIM), and Normalized Cross Correlation (NCC). These metrics evaluate the pixel-wise accuracy, perceptual quality, and structural consistency of the reconstructed frames compared to the webcam ground truth.
\section{Training Details}
\label{sec:training-details}

The present section describes the training details of the proposed facial micro-expression analysis framework.

For AU recognition, given the reduced size of our preliminary dataset, we performed Leave-One-Out Cross-Validation per subject, ensuring no data leakage.
For both image modalities, cropped frames of facial bounding boxes were resized to $64\times 64$. The SNNs were trained using the ADAM~\cite{Kingma2014AdamAM} optimizer, Mean Squared Error (MSE) as the loss function, and cosine annealing scheduler~\cite{Loshchilov2016SGDRSG} during 100 epochs. After brief hyperparameter exploration, the best found models were trained with a learning rate of 0.001, setting dropout to 0.2, and a batch size of 8 due to GPU memory constraints. 
To have a comparative baseline against a similar model, in terms of size and speed, we also trained a 3DCNN architecture based on~\cite{7410867} adapted for variable clip length. The training configuration was the same as the SNNs but with a smaller learning rate of $10^{-4}$.

For these preliminary experiments, we decided to set the slice size of the event representation to 33 ms, yielding a comparable frame rate to that of typical RGB cameras. Since event data from the Prophesee EVK4 was only available for a few subjects at the moment of writing, we decided to focus the analysis of Action Unit recognition solely on the other two cameras.
All experiments were implemented in PyTorch and ran on an Intel Xeon server equipped with two NVIDIA RTX 3080 GPUs. The hardware specification is detailed in Table~\ref{tab:server}.

\begin{table}[!h]
    \centering
    \small
    \caption{Hardware specifications of the server used to train the SNN for AU recognition. }
    \label{tab:server}
    \begin{tabular}{cccc}
    \toprule
    CPU     & 
    Max Freq. & \#Threads & Cache \\ \midrule 
    Xeon® E5-2680 v3       & 3.30 GHz        & 24 & 30MB    \\ \bottomrule \toprule
    GPU & Boost clock & Bandwidth & VRAM   \\  \midrule
    NVIDIA® RTX 3080   & 1.71 GHz & 760GB/sec    & 12GB  \\ \bottomrule
    \end{tabular}
\end{table}

For frame micro-expression reconstruction training, we used an Adam optimizer with a learning rate of $10^{-3}$, batch size of 16, and trained each model for 100 epochs. Two datasets were used: one using DVS-events as conditional input, and the other using EVK4-events. The loss function combines MSE reconstruction loss and the Kullback–Leibler divergence term from the VAE formulation.

The large volume of asynchronous events requires substantial computational resources and leads to long training times, particularly for the frame reconstruction task. To handle this, the cVAE models were trained on an NVIDIA® DGX-1™ system optimized for deep learning workloads. The hardware specifications are detailed in Table~\ref{tab:DGX-1}.

\begin{table}[!h]
    \centering
    \small
    \caption{Hardware specifications of the NVIDIA® DGX-1™ system used to train the Conditional Variational Autoencoder for event-based micro-expression frame reconstruction.}
    \label{tab:DGX-1}
    \begin{tabular}{cccc}
    \toprule
    Hardware     & Accelerator & 
    Architecture & Boost clk\\ \midrule 
    NVIDIA DGX-1 & V100        & Volta        & 1530 MHz   \\ \bottomrule \toprule
    Memory Clock & Bandwidth & VRAM & GPU   \\  \midrule
    1.75Gbit/s   & 900GB/sec       & 32GB & GV100 \\ \bottomrule
    \end{tabular}
\end{table}

\section{Results}
\label{sec:results}

This section reports the preliminary results for micro-expression recognition and intensity frame reconstruction.

\subsection{Event-based Action Unit recognition}

We evaluated the performance of the SNN model in AU recognition on our preliminary dataset, and we compared both image modalities. Our results show that the selected architectures perform significantly better when trained with event-based frames rather than RGB, as shown in Table~\ref{tab:results}. This is in line with previous findings~\cite{cultrera2024spatiotemporaltransformersactionunit, Adra2024BeyondRT} where it was shown that the event modality outperforms RGB in AU recognition using spatio-temporal models like 3D CNNs and CNNs + LSTM/Transformer. 
\begin{table}[h]
  \centering
  \small
  \begin{tabular}{ccccc}
    \toprule
    Model & \#Par. & Modal. & Acc.~(\%) & Top-3 Acc.~(\%) \\
    \midrule
    \multirow{2}{*}{SNN} & \multirow{2}{*}{4.11M} & events &\textbf{ 51.23} & \textbf{63.57} \\
    & & rgb & 23.12 & 25.57 \\
    \midrule
    \multirow{2}{*}{3DCNN} & \multirow{2}{*}{13.06M} & events & 44.89 & 56.46 \\
    & & rgb & 17.06 & 21.76 \\
    \bottomrule
  \end{tabular}
  \caption{Accuracy results for AU recognition after LOOCV for both models and modalities (webcam RGB and Davis Events).}
  \label{tab:results}
\end{table}

Additionally, the SNN outperformed the 3DCNN model by a margin of roughly 6\% in Accuracy, showing a promising direction to deepen the study of bio-inspired networks in micro-expression analysis. Nevertheless, we emphasize that these results, although encouraging, are still preliminary since the data collected lacks diversity and size; there are just seven samples for each Action Unit.

An important limitation of the experiment is the fixed time window selected to integrate events. This parameter could play a key role in the way events are interpreted, and thus it should be addressed carefully in future experiments.

\subsection{Frame micro-expression reconstruction conditioned by event frames}

\begin{figure}[htbp]
    \centering

    \subcaptionbox{}[0.48\linewidth]{%
        \includegraphics[width=\linewidth]{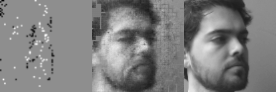}
    }
    \hfill
    \subcaptionbox{}[0.48\linewidth]{%
        \includegraphics[width=\linewidth]{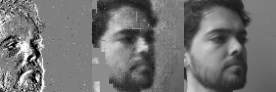}
    }
    
    \subcaptionbox{}[0.48\linewidth]{%
        \includegraphics[width=\linewidth]{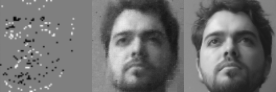}
    }
    \hfill
    \subcaptionbox{}[0.48\linewidth]{%
        \includegraphics[width=\linewidth]{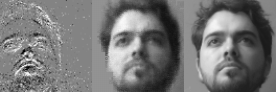}
    }
    
    \subcaptionbox{}[0.48\linewidth]{%
        \includegraphics[width=\linewidth]{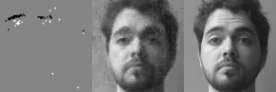}
    }
    \hfill
    \subcaptionbox{}[0.48\linewidth]{%
        \includegraphics[width=\linewidth]{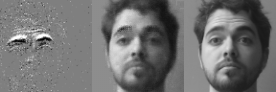}
    }

    \caption{Examples of event-based micro-expression reconstruction of intensity frames. From left to right, each image shows the input event-based frame, the reconstructed frame, and the ground truth frame. The images (a,b,c) use the DAVIS346 sensor as input and (e,d,f) use the EVK4 sensor.}
    \label{fig:reconstruction_examples}
\end{figure}

The qualitative results in Figure~\ref{fig:reconstruction_examples} show that both models attempt to reconstruct the target webcam frames from event-based inputs, preserving the general facial structure and expressions. However, differences in sharpness and noise patterns are evident between the DAVIS346 and EVK4 inputs, with EVK4-based reconstructions appearing visually smoother and closer to the ground truth.

\begin{table}[htbp]
\centering
\small
\caption{Reconstruction metrics for DAVIS346-events and EVK4-events to grayscale webcam frame.}
\begin{tabular}{lcc}
\hline
\textbf{Metric} & \textbf{DAVIS346} & \textbf{EVK4} \\
\hline
Model Params. & 289M & 289M \\
Mean MSE & 0.0034 & 0.0024 \\
Mean RMSE & 0.0566 & 0.0471 \\
Mean MAE & 0.0399 & 0.0340 \\
Mean PSNR & 25.31 dB & 26.89 dB \\
Mean SSIM & 0.7717 & 0.8513 \\
Mean NCC & 0.9613 & 0.9744 \\
\hline
\end{tabular}
\label{tab:reconstruction_metrics}
\end{table}

Quantitatively (Table~\ref{tab:reconstruction_metrics}), the EVK4-conditioned model achieves lower reconstruction errors across all metrics, with a Mean MSE of 0.002445 and a higher SSIM of 0.8513, indicating better pixel-wise accuracy and perceptual quality compared to the DAVIS346-conditioned model (SSIM = 0.7717).
Despite these differences, it is important to note that the reconstructed frames still lack the level of detail, texture consistency, and identity-preserving features required for reliable downstream tasks. Further advances in model architecture and training strategies will be necessary to improve reconstruction quality from event-based inputs.

\section{Conclusions and Future Work}
\label{sec:6_conclusions}

This work explored the spatial-temporal dynamics of facial micro-expressions through multimodal data acquisition with RGB and event cameras. We introduced a novel, preliminary dataset recorded under varying lighting conditions with multiple event-based sensors, providing a first step towards systematic evaluation of Action Units in this modality. The dataset was used to study two baseline tasks: (a) Action Unit recognition using Spiking Neural Networks trained with surrogate gradient descent, which achieved 51.23\% accuracy with event data compared to 23.12\% with RGB inputs; and (b) frame reconstruction with Conditional Variational Autoencoders, where EVK4-conditioned models outperformed DAVIS346 in terms of perceptual similarity (SSIM = 0.8513, PSNR = 26.89 dB) but still lacked detail and identity preservation.

Despite these promising results, several limitations remain. The dataset is constrained by the number of participants, resulting in gender imbalance and underrepresentation of specific age groups. Event recordings from the EVK4 were only available for a subset of subjects, reducing the consistency of cross-sensor analysis. The dataset also lacks expert-coded AU activations and activation levels, as recordings were based on elicited performances rather than certified annotations.

Future work will address these limitations by expanding the dataset with a broader participant pool, more balanced demographics, and additional recording conditions such as illumination and pose variability. Beyond dataset growth, promising directions include the design of new event-based facial representations, more advanced generative models for reconstruction, and improved methods for Action Unit recognition. Finally, multimodal fusion of RGB and event streams offers a compelling path to leverage complementary information and further advance micro-expression analysis.

\section*{Acknowledgments}
\label{sec:ack}

This work was partially funded by FONDEQUIP Project EQM170041 and the 2025 IEEE CIS Graduate Student Research Grants. 
We thank all participants who contributed to the database and the University of Chile as the venue.

{
    \small
    \bibliographystyle{ieeenat_fullname}
    \bibliography{main}
}

\end{document}